\documentclass{article}
\usepackage{ijcai24}

\usepackage{times}
\usepackage{soul}
\usepackage{url}
\usepackage{multirow}
\usepackage[hidelinks]{hyperref}
\usepackage[utf8]{inputenc}
\usepackage[small]{caption}
\usepackage{graphicx}
\usepackage{amsmath}
\usepackage{amsthm}
\usepackage{booktabs}
\usepackage{algorithm}
\usepackage{algorithmic}
\usepackage[switch]{lineno}
\usepackage{graphicx}
\usepackage{amsthm,amsmath,amssymb,amsfonts}
\usepackage{subfigure}
\usepackage{color}
\usepackage{bm}
\usepackage{multirow}
\usepackage{array}
\usepackage{rotfloat}
\def\bW{\textbf{W}}
\usepackage[table,xcdraw]{xcolor}
\newcolumntype{L}[1]{>{\raggedright\let\newline\\\arraybackslash\hspace{0pt}}m{#1}}
\newcolumntype{C}[1]{>{\centering\let\newline  \\\arraybackslash\hspace{0pt}}m{#1}}
\newcolumntype{R}[1]{>{\raggedleft\let\newline \\\arraybackslash\hspace{0pt}}m{#1}}

\title{Towards Versatile Graph Learning Approach: from the Perspective of Large Language Models}

\author{
Lanning Wei$^{1,2}$
Jun Gao $^3$
Huan Zhao$^{4}$\and
Quanming Yao$^5$
\affiliations
$^1$Institute of Computing Technology, Chinese Academy of Sciences \\
$^2$University of Chinese Academy of Sciences \\
$^3$Harbin Institute of Technology (Shenzhen) 
$^4$4Paradigm. Inc \\
$^5$Department of Electronic Engineering, Tsinghua University
\emails
\{weilanning1997,imgaojun\}@gmail.com,
zhaohuan@4paradigm.com, qyaoaa@tsinghua.edu.cn
}
\begin{document}

\maketitle

\begin{abstract}
%Recently, large language models (LLMs) have undergone rapid explorations of their potentials and applications due to the strong ability in understanding the natural languages and achieving human-like intelligence.
%For graphs that commonly used structured data in real-world scenarios, they have received attentions in exploring the feasibility of applying LLMs to address graph machine learning problems.
Graph-structured data are the commonly used and have wide application scenarios in the real world.
For these diverse applications, the vast variety of learning tasks, graph domains, and complex graph learning procedures present challenges for human experts when designing versatile graph learning approaches.
Facing these challenges, large language models (LLMs) offer a potential solution due to the extensive knowledge and the human-like intelligence.
This paper proposes a novel conceptual prototype for designing versatile graph learning methods with LLMs, with a particular focus on the ``where'' and ``how'' perspectives.
From the ``where'' perspective, we summarize four key graph learning procedures, including task definition, graph data feature engineering, model selection and optimization, deployment and serving. We then explore the application scenarios of LLMs in these procedures across a wider spectrum.
In the ``how'' perspective, we align the abilities of LLMs with the requirements of each procedure.
Finally, we point out the promising directions that could better leverage the strength of LLMs towards versatile graph learning methods.
The related source can be found at: \hyperlink{https://github.com/wei-ln/versatile-graph-learning-approaches}{https://github.com/wei-ln/versatile-graph-learning-approaches}.
\footnote{Work are done when Lanning and Jun are interns in 4Paradigm.}
\end{abstract}

\section{Introduction}

Graph-structured data are commonly used and applied in real-world applications, e.g., social networks~\cite{hamilton2017inductive}, chemistry and biomedical molecules~\cite{gilmer2017neural}.
%In the literature, efficiency and effective 
%When solving these diverse applications and data, the challenges are designing effective methods with the human expertise.
% existing methods carefully configured the procedures towards designing efficient and effective graph learning methods.
Existing methods have achieved great success in understanding and solving the single task, e.g., 
formulating the molecular property prediction as graph classification task~\cite{gilmer2017neural},  
conducting graph sampling when facing large-scale graphs~\cite{hamilton2017inductive}, designing expressive graph learning algorithms (GLAs) to extract graph structural information~\cite{min2022transformer,wu2020comprehensive}, and selecting appropriate hyper-parameters in evaluation stage~\cite{zhang2022efficient}.
%ranging from problem formulation, graph preparation, model design to tuning.
Despite their success on one single task, 
there are still challenges in graph learning that are yet to be addressed.
Firstly, the graph domains and the learning problems are largely different in real world.
Subsequently, human experts need to configure the complex procedures \cite{you2020design} individually, which brings about substantial requirements for professionalism and domain knowledge on graph-structured data.
%which is expensive and require considerable requirements on professionalism and  the domain knowledge on graph-structured data.
All these aspects lead to the challenge for human experts in handling the graph learning problems that \textit{suited for different tasks, graphs in diverse domains, and can efficiently conduct the pipeline with as few as possible assistance from humans, i.e., towards versatile graph learning methods}.
%'data from different domains, conducted with 
%effective graph learning strategies.
%\huan{Here the challenge is not that evident. Try to emphasize the challenges here, which can motivate the introduce of ``versatile'' graph learning approaches. Btw, what is the benefit/advantage of versatile graph learning approaches? As an analogy, we can clear explain what is and what is the benefit of autograph.}
%There is a considerable gap towards solving different tasks on diverse data, efficient conducting the machine learning strategies
%versatile graph learning methods 
%It indicates that there still have a long distance towards versatile graph learning methods to handling different graphs and learning tasks.
%It indicates that there is a considerable gap towards versatile graph learning methods capable of handling different graphs and learning tasks.
%they are inflexible and universal given the diverse graph learning problems in real-world~\cite{sun2023all,sun2022gppt}. 
%Then, a natural question is \textit{how to design flexible and universal techniques to handling the diverse graph learning problems}.
%The procedures to learn on graphs: problem formulation, pre-process, model design and tuning. The key challenge: designing flexible and universal techniques to learning on graphs?

%The large language models (LLMs) have advantages in handling diverse tasks through natural languages, and then the flexible and universal techniques motivate us to employing them and handling the above problem.
Large language models (LLMs) are treated as the key point in designing versatile graph learning algorithms due to the maintained knowledge and human-like intelligence.
%why they can solve the above problem.
LLMs, refer to the large-sized (billions-level) pre-trained language models (PLM) in general,  
have undergone a rapid succession of breakthroughs in recent years~\cite{zhao2023survey}.
By pre-training with comprehensive text data and prompt-tuning on downstream tasks, 
%LLMs have improved capacity on the diverse downstream tasks~\cite{zhao2023survey}.
LLMs pose a vast store of knowledge and show the ability in achieving the human-level decision-making ability~\cite{zhao2023survey,wang2023survey}. 
For instance, LLMs can reason on mathematical problems with chain-of-thoughts~\cite{wei2022chain}, span different tasks and achieve superior performance~\cite{mialon2023gaia}, solve computer vision tasks from task planning to algorithm selection and execution like human experts~\cite{shen2023hugginggpt}.
%\footnote{+wln+ it is correct to say this? maintained knowledge}
% and then they could handling the different task
Based on these abilities, LLMs show promising potential to serve as Artificial General Intelligence (AGI)~\cite{ge2023openagi} and general research assistant~\cite{huang2023benchmarking}.
Therefore, it is natural to directly use or draw on the successful experience of LLMs when constructing versatile graph learning methods.

\begin{figure*}[ht]
	\centering
	\includegraphics[width=0.95\linewidth]{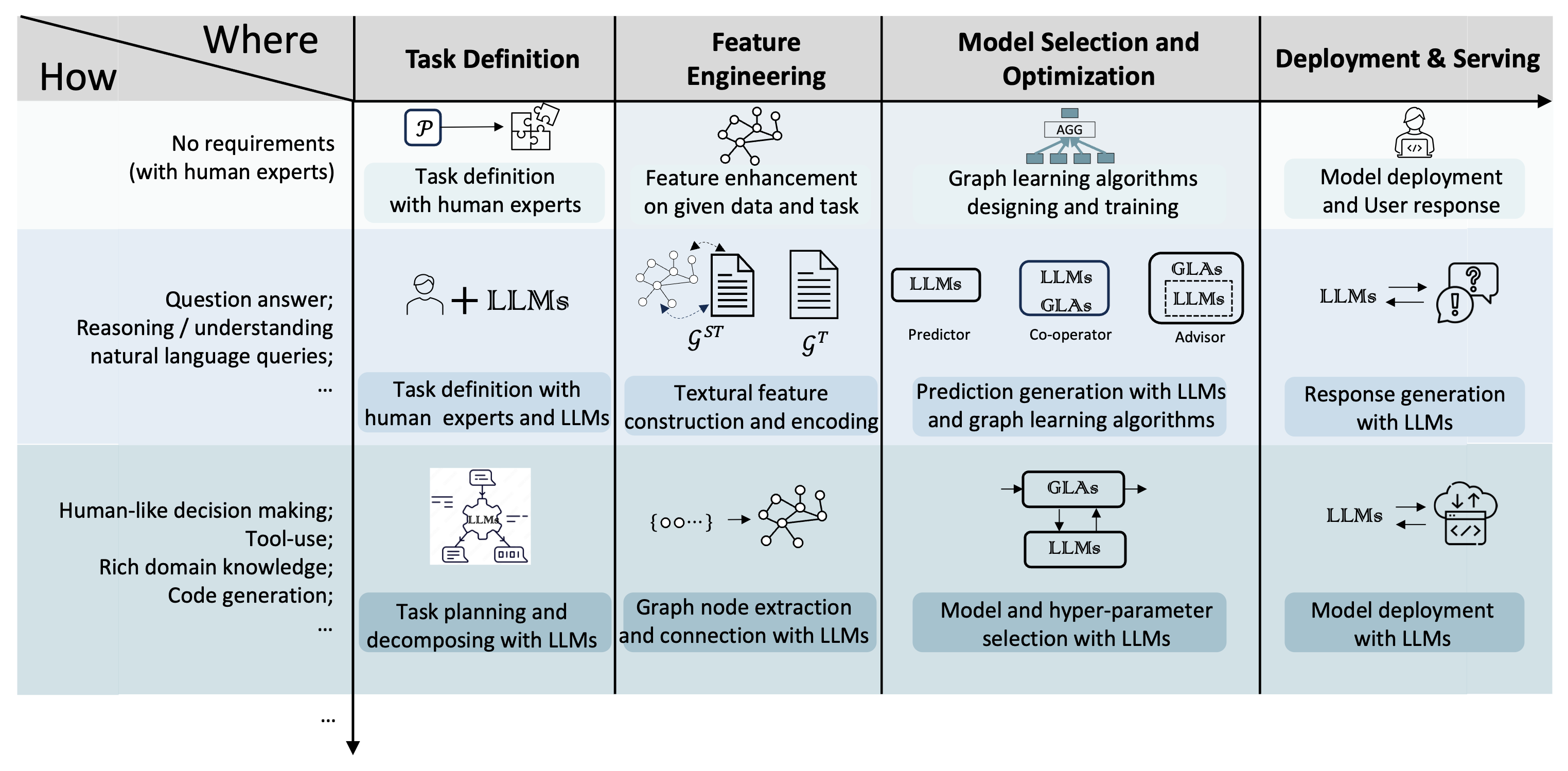}
	\caption{The conceptual prototype of versatile
		graph learning methods joint with LLMs. 
		LLMs can be used in sequential graph learning procedures in these columns, with increased requirements for LLMs in different rows for each procedure.
		The rows can be further developed along with the exploration of different abilities of LLMs.
	}
	\label{fig-overview}
\end{figure*}

%pathway to AGI? research assistant.
%Motivated by the above strengths of LLMs,
%there is a growing interest in employing LLMs when constructing graph learning methods
The numerous advantages of LLMs have triggered an increasing interest in using them for graph learning problems.
%in handling the diverse graph-structured data and graph learning problems with LLMs~
\cite{li2023survey,guo2023gpt4graph,shen2023hugginggpt,zhang2023automl}. 
For instance, through harnessing the preserved knowledge , different graph learning tasks across various domains can be unified into textual descriptions, which subsequently employ LLMs, such as ChatGPT, to acquire predictions
%and then employ LLMs like ChatGPT to obtaining the predictions
%with the LLM-empowered graph encoding and prompt engineering
~\cite{liu2023one,fatemi2023talk}. 
LLMs enhance the textural-attributed graphs (TAGs) which subsequently employ LLMs, such as ChatGPT, to acquire predictions~\cite{he2023explanations,guo2023gpt4graph,chen2023exploring}.
These methods have catalyzed impressive progress in designing effective techniques for different graph learning problems.
%\footnote{+wln+ select representative methods and then revise based on the categorized in the following.}

% or predictor to encode the textual information and obtain better performance on different tasks~\cite{li2023survey,pan2023integrating,chen2023exploring}.

%Beyond language models to encode the textual information merely~\cite{he2023explanations}, 

%Beyond these methods, the potential of LLMs have not been fully unleashed considering the 
%have billions of parameters and show 
%promising ability to serve as Artificial General Intelligence (AGI)~\cite{ge2023openagi} and general research assistant~\cite{huang2023benchmarking}. 
%Nevertheless, existing methods on graph learning have few explorations and cannot unleashing the powers of LLMs.
%To incorporating LLMs in graph learning procedures from this new perspective, the key insights are the identification of:

With such a development of graph learning methods joint with LLMs, it is necessary to explore the systematical way towards versatile graph learning methods using LLMs.
In this paper, we conduct a comprehensive review on existing methods and develop conceptual prototype
by aligning the advantages and strong abilities of LLMs with the different requirements maintained in versatile graph learning methods.
More specific, the key insights are the identification of:

\noindent\textbf{Where can be used}.
%Existing methods mainly use LLMs in data construction and model selection, while ignore the other graph learning procedures~\cite{chen2023exploring,li2023survey,wei2023unleashing}. 
%LLMs have advantages in achieving human-like decision making, which enables to decompose and complete the complex graph learning tasks~\cite{wang2023survey,zhao2023survey}.
%Human experts paid sufficient and configured? when facing different tasks
Different requirements for LLMs are arise when human experts manage the graph learning pipeline.
%It indicates the potential that LLMs can be applied to the different procedures like human experts~\cite{wang2023survey,shen2023hugginggpt}. 
%Given its superiority,
Towards versatile graph learning approach,
it is a vital to examine \textit{the feasibility of LLMs in diverse graph learning procedures} to fully unleash the versatile abilities of LLMs.
%Therefore, 
%as shown in Fig.~\ref{fig-overview}, 
Consequently,
we summarize four key procedures based on the requirements according to a standard machine learning pipeline, i.e., task definition, feature engineering, model selection and optimization, deployment and serving, as shown in Fig.~\ref{fig-overview}.
%and then 
Based on these procedures, we can explore a wider spectrum of application scenarios when employing LLMs, thereby enhancing the versatility of existing graph learning methods.
%\footnote{+wln+ update this sentence. it looks strange}
%from the learning perspective, LLMs have advantages in designing understanding the different tasks
%. problem formulation and make human-like decisions. which could serve as the Graph learning agent balabala.
%\footnote{+wln+ unify the name of graph learning procedures.}
%principle: involved in more procedures.

%Beyond that, graph nas, agent graph representation learning.
%serve as enhancer or preditor. When applying the human-level intelligence.

\noindent\textbf{How to use}.
%LLMs can be treated as merely models, language models or human-like decision maker due to its strong ability in encoding, understand and decision-making abilities, respectively.
%Existing methods have explored the LLMs that serves as model to enhancing graph embedding and language models that encoding the textual information~\cite{liu2023one,guo2023gpt4graph}. 
%Given the potential of LLMs and the diverse different requirements of graph learning procedures,  
It is important to align the capabilities of LLMs with the requirements of the four procedures.
%It is natural that different procedures have totally different targets and requirements when leverage the strengths of LLMs.
%Therefore, we provide a detailed analysis on existing graph learning methods to summarize the match guideline.
%on the relations between the employment of LLMs and the existing methods.
%how LLMs with different abilities are involved in these procedure.
%on the how graph learning methods integrated with LLMs based on different abilities.
%LLMs with different abilities used in these procedures.
%As illustrated in Fig.~\ref{fig-overview}, 
%three levels of increased requirements for LLMs are summarize
We summarize three levels of increased requirements for LLMs, depicted as three rows in Fig.~\ref{fig-overview},
% which represent the increased requirements of LLMs', 
and then provide a detailed analysis of existing methods.
The human-designed methods, represented in the first row, place no requirements on LLMs, 
The methods situated in the second row emphasize the fundamental ability of LLMs to understand, encode, and reason out the natural language queries.
%i.e., collaborating with human experts and operating independently as an agent.
The methods in the third row require a higher level of human-level decision-making ability and the rich domain-specific knowledge from LLMs, similar to human experts. 
%\footnote{+wln+ may be not exact? how to evaluate the intelligence of LLMs? or how to describe the role act as agent and act as a language model?}
The varying capabilities of LLMs encourage us to investigate their potential applications in each procedure, prompting the development of versatile graph learning methods.
%
%However, there is no guideline to match the LLM techniques with the procedures requirements.
%The clarification will help identifying how to combining LLMs in graph learning and unleashing the power of LLMs.
%Therefore, we provide
%exploring the potential in each procedures and all pipelines.
%principle: model/LM/LLM. in different procedures. they have different advantages and disadvantages in current years. used in the long-range future and used in current have different understandings. why then cannot be used in current stage.

It is significant to highlight the proposed \textbf{conceptual prototype} in Fig.~\ref{fig-overview} when designing versatile graph learning methods based on LLMs.
The ``Where'' perspective is spanned along with four key procedures in graph learning pipeline, and the ``How'' perspective is organized based on the abilities of LLMs in different levels. 
Then, the graph learning methods joint with LLMs can be designed by choosing the desired ability in each procedure, which is detailed shown in Table~\ref{tb-overview}.
Based on these two perspectives, we provide \textbf{comprehensive overview} of the graph learning methods jointed with LLMs, and emphasize the potential for a broad exploration spectrum and the usability of LLMs' various abilities towards versatile graph learning approaches.
%detailed analysis, comparisons and discussions to guide the match between LLM techniques and the construction of versatile graph learning methods.
Finally, we suggest promising \textbf{future directions} on the basis of under-explored ability of LLMs and the property of graph-structured data, i.e., the effectiveness in understanding the graph structure, large graph foundation model, and universal graph learning agents.

%To summarize, the contributions of this paper are as follows:
%\begin{itemize}
%	\item \textbf{Comprehensive Overview:} We give a comprehensive overview on the graph learning methods jointed with the LLMs, and emphasize the potential for a broader exploration spectrum and the usability of LLMs' various abilities towards versatile graph learning approaches.
%	\item \textbf{Systematic Taxonomy:} We provide a systematic taxonomy of the graph learning methods jointed with the LLMs based on 
%	``Where can be used'' and ``How to use''. 	
%	``Where'' perspective is spanned along with four key procedures in graph learning pipeline, and ``How'' perspective is organized based on the different roles that LLM played. 
%	In each category, we provide detailed analysis, comparisons and discussions to guide the match between LLM techniques and the construction of versatile graph learning methods.
%%		from the perspective of graph learning 
%%	 along with the graph learning procedures and the involvements of LLMs.
%	\item \textbf{Future Directions:}We suggest the promising future directions on the basis of unexplored ability of LLMs and the property of graph-structured data, i.e., the effectiveness in understanding the graph structure, large graph foundation model, and universal graph learning agents.
%	\footnote{+wln+ update after re-write the future directions.}
%\end{itemize}

\section{Overview}
\label{sec-preliminary}

\subsection{Machine Learning on Graphs}

%\footnote{+qm+ your taxonomy is based on this part, 
%	a better way to sell yourself is to put this as motivation to your taxonomy.
%	No need to put this as ``related works / preliminary''.}
The graph-structured data is presented as $\mathcal{G}=(\mathcal{V}, \mathcal{E})$, 
where $\mathcal{V}$ and $\mathcal{E}$ are nodes and edges. $\textbf{A} \in \mathbb{R}^{|\mathcal{V}| \times |\mathcal{V}|}$ is the adjacency matrix of this graph, 
and $\textbf{H}\in \mathbb{R}^{|\mathcal{V}| \times d}$ is the feature matrix.  
%To use the graph machine learning techniques 
The graph learning procedures generate the results $\mathcal{R}$ of graphs learning problem $\mathcal{P}$ on graph $\mathcal{G}$.
%i.e., $f: (\mathcal{G}, \mathcal{P})\rightarrow \mathcal{R}$.
%\footnote{+wln+ the notations may not use in the following paragraph. upd this part.}

Considering the considerable efforts expended by human experts to solve these diverse graphs and applications, we summarize the key points in graph learning pipeline
%The graph learning procedures can be conducted with human experts step by step 
as shown in Fig.~\ref{fig-overview}.
%\footnote{+wln+ introduce the inner circle}

\begin{itemize}
\item 
%\footnote{+qm+ is it better say ``Graph Learning Task Construction / Definition''?
%	Same for the title of Section~3.1.
%	``Problem Definition'' looks like you are introducing a formal math def to a problem.\checkmark}
%Problem Definition. 
Task definition.
Experts formulate the problems $\mathcal{P}$ into specific graph learning tasks $\mathcal{T}$, 
e.g., the predictions on nodes, edges, sub-graphs or graphs.

\item Feature engineering. Experts select, combine and transform features towards better-performed performance by themselves~\cite{liu2022taxonomy}. 
For instance, different frequency bands are designed to eliminate certain graph Fourier modes. 
Besides, it can be achieved automatically with AutoML~\cite{zheng2023automl}, 
which involves selecting appropriate feature sets and combining them in a data-driven manner.

\item Model selection and optimization. 
Experts select and train architectures $\mathcal{A}(\bW^*)$, e.g., Graph Neural Networks~\cite{wu2020comprehensive} or Graph Transformers~\cite{min2022transformer}, to address the graph learning problem. Here, $\mathcal{A}$ represents the selected architecture with the trained parameter $\bW^*$. 
In addition to designing and tuning architectures with human experts, other strategies including graph neural architecture search (NAS) and hyper-parameter tuning methods are explored, which aim to automated conducting this procedure and obtaining efficient and effective solutions~\cite{zhang2021automated,zhang2022efficient}.

\item Deployment and serving. 
Experts deploy models and then generate the results $\mathcal{R}$ to serve for users and graph learning problem $\mathcal{P}$, which are achieved by experts themselves.
\end{itemize}

\subsection{Conceptual Prototype}
%\footnote{+qm+ should this part be outside sec~2?
%	this is what you proposed, should not be ``preliminary''?}
In this paper, we propose 
%\huan{Too trivial to use this word. Should use ``propose/develop'' to strengthen the importance of this prototype.} 
a conceptual prototype when designing versatile graph learning methods jointed with LLMs, 
%give taxonomies of existing graph learning methods 
following the development of graph learning procedures and the different abilities of LLMs as illustrated in Fig.~\ref{fig-overview}.
%\footnote{+wln+ 
%	1. maybe we can categorize the pipeline into three parts: problem definition / data (collection, feature engineering) / model (selection/optimization/evaluation)  following the \cite{yao2018taking}.  
%	2. 
%}

\noindent\textbf{``Where can be used''}
%Based on four key graph learning procedures, i.e., four ordered sectors in Fig.~\ref{fig-overview},
%we summarized the common requirements of each procedure, on top of which we explore the potential application scenarios of LLMs in each procedure, 
%leading to a broader spectrum to apply the LLMs towards versatile graph learning methods.
We analyze and summarize the common requirements in each key graph learning procedure, which are represented as four columns in Fig. \ref{fig-overview}. 
This analysis led us to explore the potential application scenarios of LLMs in each procedure, which could broaden the spectrum of potential applications of LLMs in various graph learning methods.
The results show that existing methods are extensively used in the feature engineering and model selection procedures, attributing to the need for enhancing features and representations of textual data. 

%The results identify that existing methods are extensively explored in the feature engineering and model selection procedures
%due to the requirements in enhancing the features and representations on the textual data.
% due to the ability of  encoding and understanding textual information potentially maintained in graph.
%For the problem definition, deployment and response procedures which only achieved by expplan and determined the 

\noindent\textbf{``How to use"}
Considering the different requirements in each procedure, 
%three concentric circles represent the different involvements of LLMs, 
%from human experts (without LLMs), (LLM-assist)  to agent (LLM-based)
%along with the y axis, 
we explore the feasibility of using LLMs according to a hierarchy of requirements for LLMs' abilities. 
%following the different levels of requirements for LLMs' abilities 
This hierarchy is visualized as different rows in Fig.~\ref{fig-overview},
%as the concentric circles illustrated in Fig.~\ref{fig-overview}, 
i.e., 
%conducting procedures with human experts (without LLMs)
no requirements for the use of LLMs (with human experts merely); requiring the fundamental ability in understanding and reasoning the natural languages; and requisition of advanced human-like intelligence and domain-specific knowledge as human experts.
%By justifying the match guideline between procedure requirements and LLM abilities, we could explore the versatile applications in graph learning methods.
%\footnote{LLM-embedded?}
%which serve as models and are used in the procedures of pipeline.
%As shown in the middle circle, 
The majority methods, represented in the second row, employ LLMs to assist the human experts due to their proficiency in question answer, understanding and encoding the natural language queries.
Conversely, a minority methods, as illustrated in the final row, exploit the potential of LLMs in planning, decomposing and completing tasks with themselves due to the advanced ability in achieving human-like intelligence.
\textit{It is necessary to note that the potential abilities of LLMs are still being explored and the conceptual prototype represented in the figure may be extended to include more rows as the development of LLMs progresses.}
%\footnote{+wln+ need to update}

%\footnote{+wln+ in each axis, introduce how/why we group methods, and provide one sentence observation. }

\subsection{Comparisons with Contemporaneous Surveys}
There exists several contemporaneous surveys that explore how to use LLMs in feature engineering and model selection procedures.
To be specific, \cite{li2023survey,jin2023large} categorized the methods based on the role that LLM played, 
and \cite{zhang2023large,liu2023towards} categorized the methods from the perspective of graph foundation models, 
ranging from backbone architecture construction and pre-training to adaption.
Compared with these methods, this paper considers the full graph learning procedures and the versatile abilities of LLMs.
%potential of LLMs as an assistant.
Firstly, this paper takes into account the complete graph learning processes,  e.g., 
%further considers 
the task definition, deployment and serving procedures, 
%which makeup the full graph learning pipeline when using LLMs~\cite{yao2018taking} and 
that provide a wider spectrum of application scenarios for LLMs.
Besides, this paper further considers the ability of LLMs in achieving human-like intelligence.
It indicates that LLMs can serve as core orchestrators of autonomous agents, representing a new trend of LLMs~\cite{wang2023survey}.
%which is consistent with the recent research topic in the world~\cite{he2023explanations}.
This suggestion aligns with recent global research topics~\cite{he2023explanations}.

\section{Graph Learning with LLMs}
In this section, we overview the existing methods 
%from two perspectives, i.e., introducing how LLMs are utilized in the graph learning procedures step by step.
following the sequential graph learning pipeline, and then the methods used in each procedure are introduced following the involvement of LLMs.
%\footnote{+wln+ emphasize the match guideline between requirements and LLMs techniques? we have mentioned them in the first paragraph.}
%When LLMs used machine learning problems on graphs, we first introduce the procedures related to problem definition, deployment and response, reflecting the interaction among users. Subsequently, we present the processes involved in data preparation and algorithm selection, which are heavily associated with graph-structured data.

%In this section, we introduce each procedure in graph learning and how LLMs is combined with them.

%\footnote{+wln+ in each subsection. Introduce the key insight/challenge of the designed method, and then introduce why LLMs can achieve this. Introduce the LLM+graph methods. Give summary and suggest the things can be explored based on the LLMs potential or things cannot do due to its disadvantages, how to meet this gap. }
\begin{table*}[ht]
	\centering
	\tiny
	\setlength\tabcolsep{1pt}
	\caption{A summary of graph learning methods jointed with LLMs under the proposed conceptual prototype. represents cells that were created by human experts and do not have any relation to the following procedures that use LLMs. 
%		The colors represent the requirements of LLMs which is consist with Fig.~\ref{fig-overview}	
}
	\begin{tabular}{l|L{40pt}|l|L{70pt}|L{40pt}|L{40pt}|L{70pt}|L{40pt}|L{50pt}}
		\hline
		\multicolumn{1}{c|}{\multirow{2}{*}{Methods}} & \multicolumn{1}{c|}{\multirow{2}{*}{\begin{tabular}[c]{@{}c@{}}Task\\ Definition\end{tabular}}} & \multicolumn{3}{c|}{Feature Engineering} & \multicolumn{3}{c|}{Model Selection and Optimization} & \multicolumn{1}{c}{\multirow{2}{*}{\begin{tabular}[c]{@{}c@{}}Deployment \&\\ Serving\end{tabular}}} \\ \cline{3-8}
		\multicolumn{1}{c|}{} & \multicolumn{1}{c|}{} & \multicolumn{1}{c|}{Taxonomy} & \multicolumn{1}{c|}{Description} & \multicolumn{1}{c|}{LLMs} & \multicolumn{1}{c|}{Taxonomy} & \multicolumn{1}{c|}{Description} & \multicolumn{1}{c|}{LLMs} & \multicolumn{1}{c}{} \\ \hline
		Graphtext~\cite{zhao2023graphtext} &  -  &   $\mathcal{G}^T$ &  The features and node labels of k- hop subgraphs &  -  &  Predictor &  predicting labels of centric node &  ChatGPT / GPT-4 &  Using predictions as response \\ \hline
		NLGraph~\cite{wang2023can} &  Formulating graph reasoning tasks with human experts & $\mathcal{G}^T$ & Describing the graphs, nodes, edges and learning tasks. &  -  &  Predictor &  Predicting the statistics of graphs. &  GPT-3.5 / GPT4 &  Using predictions as response  \\ \hline
		GPT4Graph~\cite{guo2023gpt4graph} &  -  &  $\mathcal{G}^T$ &  self-prompting to describe the graphs. &  InstructGPT- 3 &  Predictor &  Predicting the statistics and Semantic labels of graphs. &  InstructGPT-3 &  Using predictions as response \\ \hline
		TLG~\cite{fatemi2023talk} &  Formulating graph reasoning tasks with human experts &  $\mathcal{G}^T$ &  Describing the graphs, nodes, edges and learning tasks. &  -  &  Predictor &  Predicting the statistics of graphs. &  PaLM 62B &  Using predictions as response  \\ \hline
		LLM4Mol~\cite{qian2023can} &  -  & $\mathcal{G}^T$ &  Combining SMILES text and explanations of functional groups, chemical properties, and potential pharmaceutical applications &  ChatGPT & Predictor & Fine-tune a pre-trained language model on various molecule related downstream tasks & RoBERTa &  - \\ \hline
		OFA~\cite{liu2023one} &  -  &  $\mathcal{G}^{ST}$ &  Explanations on $\mathcal{V}^T$ and $\mathcal{E}^T$ &  ChatGPT &  GLA-centric &  co-trained LLMs and R-GCNs &  Sentence-bert &  -  \\ \hline
		\cite{chen2023exploring} &  -  &  $\mathcal{G}^{ST}$  &  Generating and encoding textual information of nodes &  Open-source LLMs Deberta / LLaMA &  GLA-centric / Alignment-based / Predictor  &  Combining GCN/GAT/RevGAT with Sentence-BERT/ Deberta; using Deberta as predictor  &  Combining GCN/GAT/RevGAT with Sentence-BERT/ Deberta; using Deberta as predictor  &  -   \\ \hline
		%		TAPE~\cite{he2023explanations} &  -  &  $\mathcal{G}^{ST}$ &  Explanations on $\mathcal{V}^T$ Predictions, encoding &  GPT3.5, fine-tuning Deberta &  -  &  Training RevGAT with enriched features &  Deberta / LLaMA / SentenceBERT  &  -  \\ \hline
		GRAD~\cite{mavromatis2023train} &  -  & $\mathcal{G}^{ST}$ &  Using the raw texts in graphs &  -  &  Alignment-based &  co-trained GNN teacher and LLM student &  BERT &  -  \\ \hline
		GraphGPT~\cite{tang2023graphgpt} &  -  &  $\mathcal{G}^{ST}$ &  Using the raw texts in graphs &  -  &  Alignment-based &  Applying contrastive alignment objective on the predictions of  pre-trained GNNs and LLMs &  Bert &  -  \\ \hline
		GraphPrompt~\cite{liu2023graphprompt} &  -  & $\mathcal{G}^S$ &  Sampling and unifying data and tasks &  -  &  GFMs &  Pre-train and prompt on downstream tasks &  -  & -   \\ \hline
		All in One~\cite{sun2022gppt}&  Unified with graph-level tasks  &  $\mathcal{G}^S$ &  Constructing prompt graphs &  -  &  GFMs &  Pre-trained with multi-task meta-earning and prompted on downstream tasks &  -  & -   \\ \hline
		Instruction2GL~\cite{wei2023unleashing} &   Formulating tasks with ChatGPT-3.5 &   $\mathcal{G}^S$ &   Slecting engineering strategy following user instructions &   ChatGPT-3.5 &   LLM-assisted &   Configuring AutoML and &   ChatGPT-3.5 &   Configuring codes and generating response with ChatGPT and the agents  \\ \hline
		GPT4GNAS~\cite{wang2023graph} &  -  &  $\mathcal{G}^S$ &  -  &  -  &  LLM-assisted &   Selecting and evaluating GNNs on AutoML with LLMs. &  GPT-4 &  - \\ \hline
	\end{tabular}
	\label{tb-overview}
\end{table*}

\subsection{Task Definition}
Task definition transforms the learning problems in real-world applications into possible solvable tasks with current machine learning techniques. 
It is the first step to addressing the machine learning problems and can only be effectively done through human expertise.
%Understanding the issues and the current techniques is the key challenge in this procedure. 
LLMs have advantages in understanding the different applications in real-world, and they have the ability to assist the complex task planning, decomposition and completion
% \huan{to do what? Try to summarize what specific things the RA can do.}
%been trained with the machine learning techniques in recent years
~\cite{mialon2023gaia,huang2023benchmarking}. 
%and these methods have been applied in planning the complex AI tasks in computer vision~\cite{shen2023hugginggpt}.
%\footnote{+wln+ add ref in this paragraph.}
%

Efforts have been made by combing LLMs with other modalities and graph-structured data.
In computer vision, HuggingGPT~\cite{shen2023hugginggpt} used LLMs to understand the complex computer vision tasks and complete them sequentially with the help of LLMs, which can be achieved due to the human-like decision-making capability of LLMs.
%What are special in graph learning? understanding graph learning problems? 
%\footnote{+wln+ examples in solving the NLP tasks? compared with images/NL?}
However, graph-structured data, which stems from different domains, presents a unique challenge. The learning tasks related to graph-structured data are in varied formats, necessitating diverse configurations for individual tasks~\cite{liu2023towards}.
%, which bring difficulties to understand and formulate the problems. 
In the literature, NLGraph~\cite{wang2023can} and GPT4Graph~\cite{guo2023gpt4graph} evaluated LLMs in understanding and formulating the graph reasoning tasks over different graph datasets. Instruction2GL~\cite{wei2023unleashing} proposed a LLM-based planning agent and then mapped the users' instructions into different graph learning tasks. 
Given the experiments of these methods over different tasks, the feasibility of LLMs in planning and understanding graph learning tasks is obvious.
%\footnote{+wln+ LLMs could understanding but may not addressing the problem.}

In conclusion, defining the learning tasks is the first step when facing the real-world applications on graph-structured data. 
It can be solved by LLMs with the ability in understanding the natural languages and the maintained domain knowledge over graphs.
With these abilities, LLMs bring more convenient interaction manner with users and could alleviate the stress over human experts. 
It indicates the potential of LLMs to define, planning, decomposing the complex graph learning tasks.
%serve as graph learning assistant.
%, while cannot be achieved with general LMs. 
% HCI
%Furthermore, it would change the interaction manner when learning on these graphs and applications.

\subsection{Feature Engineering}
\label{sec-feature}
Graph feature engineering is responsible for combing or transforming data to more effective features which are built based on the original graph-structured data.
%that will be used in the following algorithms $\mathcal{A}(\cdot)$. 
%The challenges are choosing strategies to prepare the graphs.
LLMs have the ability to generate descriptions of the graph using natural language, and can also encode text into an embedding space, serving as a useful tool in this regard. By introducing this new modality to graphs, an abundance of textual information can be incorporated, which may lead to a significant improvement in the effectiveness of the algorithm.
%\footnote{+wln+ potential benefits from the }

\begin{figure}[ht]
	\centering
	\includegraphics[width=0.95\linewidth]{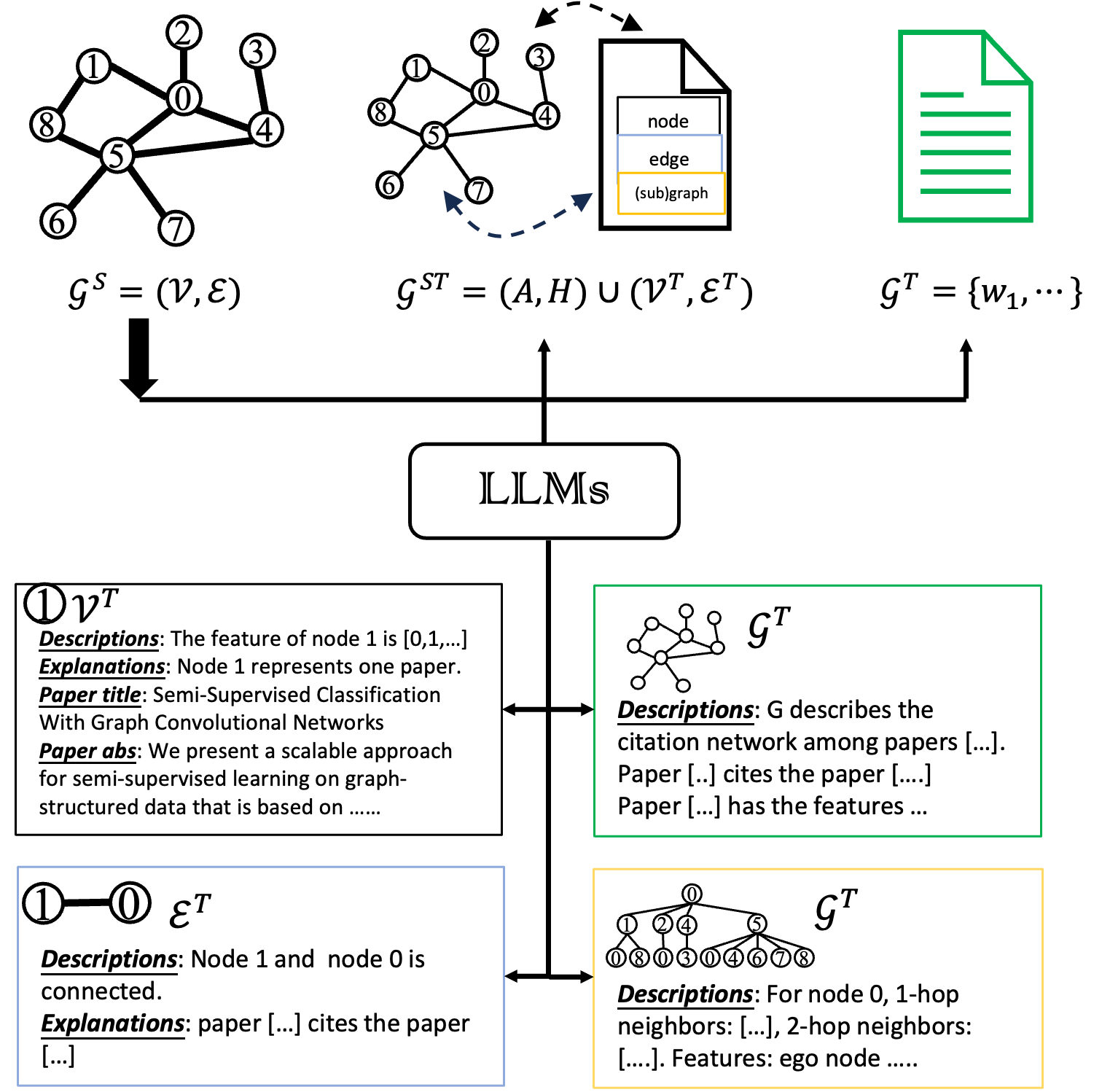}
	\caption{Illustrations of graph data feature engineering strategies that jointed with LLMs.}
	\label{fig-data-collection}
\end{figure}

%\begin{definition}[Structured Graph]
%	\label{def_gs}
%	A structured graph $\mathcal{G}^{S}$ is defined as a structure consisting of nodes $\mathcal{V}$ and edges $\mathcal{E}$.
%\end{definition}
%
%\begin{definition}[Textual Graph]
%	\label{def_gt}
%	A textual graph $\mathcal{G}^{T}$ is defined as a word sequence $\{w_1, \cdots\}$ which describes the nodes and edges in graph.
%\end{definition}
%
%\begin{definition}[Structure-text Graph]
%	\label{def_gst}
%	A structure-text graph $\mathcal{G}^{ST}$ is defined as a graph combing structures $\mathcal{G}^S$ and textual description $\{w_1, \cdots\}$ of nodes $\mathcal{V}^T$, edges $\mathcal{E}^T$, or the graph $\mathcal{G}^T$.
%\end{definition}

%\footnote{+wln+ add figs to analysis these methods.}
%Firstly, graphs can be constructed with LLMs. AutoKG~\cite{chen2023autokg} extracted the entities from knowledge bases with LLMs, which could handle complex relation dynamics due to the strong ability in understanding the textual information.
\subsubsection{Textual Feature Construction}
%The majority of existing methods tend to enhance the graphs with textual information.
As shown in Fig.~\ref{fig-data-collection}, two types of engineering strategies are designed to combine textual information
%text-enhanced graphs are obtained
based on the graph-structured data.
In this paper, for the clear justification, we denote the \textbf{structured graph} as $\mathcal{G}^S=(\mathcal{V},\mathcal{E})$,
and the \textbf{textual graph} $\mathcal{G}^T$ as a word sequence $\{w_1, \cdots\}$ which describes the nodes and edges in graph.
The \textbf{structured-textual graphs} $\mathcal{G}^{ST}$ is defined as the combination of structures $\mathcal{G}^S$ and textual description $\{w_1, \cdots\}$ of nodes $\mathcal{V}^T$, edges $\mathcal{E}^T$, or the graph $\mathcal{G}^T$.
%\footnote{We re-denoted the general graphs as $\mathcal{G}^S$ for clearer claim}.

Firstly, for structured-textual graph $\mathcal{G}^{ST}$,
%the additional textual information of nodes, edges or graphs are attached on original graph $G$~\cite{he2023explanations}, and then 
%both textual and structure information are incorporated in graph, i.e., \textbf{structure-text graph} $\mathcal{G}^{ST}=(\mathcal{V}, \mathcal{E}) \cup (\mathcal{V}^T, \mathcal{E}^T)$.
%For graphs with rich textual information, 
the textual descriptions and explanations of nodes, edges or sub-graphs can be generated by LLMs and used in the following procedures to improve the effectiveness~\cite{chen2023exploring}.
% e.g., social networks, citations graphs and molecules, 
To be specific, Chen et al.~\cite{chen2023exploring} highlighted the text attributes for nodes, GraphGPT~\cite{tang2023graphgpt} described the graph structures with natural languages,  TAPE~\cite{he2023explanations} used GPT-3.5 to generate the node descriptions, prediction as well as the explanations to enrich the text information; OFA~\cite{liu2023one} further provided the descriptions of edges and learning tasks.
%The representative methods are \cite{chen2023exploring,tang2023graphgpt,liu2023one,he2023explanations}, in which the explanations of node (edge) types and the descriptions of graph structures are generated by LLMs, e.g., the paper title and abstracts for citation networks which describe the citation relationship between papers. 
%
%the nodes are papers in the citation networks, and the abstract and introductions of the paper can be extracted~\cite{chen2023exploring,tang2023graphgpt,liu2023one}; 
%and the edge types of the citation graphs and molecule graphs can be described with different information~\cite{liu2023one}.
%the edges types and the graphs~\cite{liu2023one,}
%
For textural graph $\mathcal{G}^{T}$, 
%Apart from that, 
existing methods use the textual descriptions to replace the general structures.
%, i.e., , representing the LM-based algorithms can be directly employed for these data.
%\footnote{+wln+ how to describe the text data? how to describe the word sequence}
The representative method GraphText~\cite{zhao2023graphtext} sampled k-hop ego-subgraphs and then describe the labels and features of nodes in each hop. TLG~\cite{fatemi2023talk} further provided 11 types of text description functions.
LLM4Mol~\cite{qian2023can} describe the molecules with SMILES text, and the explanations of functional groups, chemical properties, and potential pharmaceutical applications generated by ChatGPT.
%Besides, \cite{wang2023can} provide different benmarks 
%\footnote{+wln+ these are general graph reasoning problem, rather than graph learning problems. Maybe these methods should not be added here.}

%OFA~\cite{liu2023one} describe the nodes, edges and tasks with LLMs and can be applied on all nodes.
%Apart from that, graphs $G$ can be constructed with LLMs. AutoKG~\cite{chen2023autokg} extracted the entities from knowledge bases with LLMs, which could handle complex relation dynamics due to the strong ability in understanding the textual information.

In conclusion, existing methods enhance the textual information with the help of LLMs, the enhancement 
%use LLMs in generating diverse textual information to enhancing the graphs, 
include but not limited to the descriptions on nodes and connection relationship and explanations on the rich domain information maintained in nodes, edges or graphs.
It is achieved with the abilities of answering the questions of users based on the vast knowledge maintained in LLMs.
%\footnote{+wln+ maybe we should emphasize the ability of question answer, since the prompt / enhancement queries are provided by users.}
%are treated as language models to assist the graph construction. 
%Different textual statements of graphs have significant influence on the performance of LLMs~\cite{fatemi2023talk}, and these stills need to be explored on effectively describing the graphs and efficiently constructing the graphs.

%\subsection{Feature engineering}
\subsubsection{Textual Feature Encoding}
%Based on the above text information in graph $\mathcal{G}^{TA}$ and $\mathcal{G}^T$, different open-source language models are adopted to 
%
Existing methods employ open-source LMs or LLMs to obtaining the text embedding
%of nodes~\cite{tang2023graphgpt}, edges~\cite{liu2023one}, and graphs~\cite{qian2023can} 
with vanilla Transformer~\cite{vaswani2017attention}, Sentence-BERT~\cite{reimers2019sentence}, RoBERTa~\cite{liu2019roberta} or PaLM~\cite{chowdhery2023palm}. 
These models can be frozen to obtaining the embedding directly, or co-trained with the graph learning algorithms which will be introduced in the following.
%Given the stronger ability of LLMs in encoding the text information, these methods can be replaced with open-sourced LLMs in the future work to further enhancing the effectiveness, e.g., LLaMA- 7B/70B.

\subsubsection{Summary and Discussion}
In conclusion, existing methods explore the feasibility of LLMs in generating novel text modality for graphs.
As illustrated in Fig.~\ref{fig-overview}, by combing the expertise of human experts with the versatile explore of LLMs, these methods construct and encode the textural information, on top of which the performance improvements of graph learning methods can be expected.
Apart from these methods, AutoKG~\cite{chen2023autokg} adopted LLMs to construct the entities of knowledge graphs based on natural languages, thereby empowering decisions akin to human expertise.
%with LLM-based agent, which beyond this procedure.
%\footnote{+wln+ the graph constructio}
%which is benefit for graphs with rich text information. 
Despite the success of these methods, there is a notable absence of systematic comparisons and discussions about effective construction of textual components in the graph, especially for text-sparse graphs such as traffic and power transmission graphs~\cite{jin2023large}, graph with signal or image features that are challenging to describe using natural languages.
%\footnote{+wln+ re-write, add structure description and encoding. Combing with the future work.}
%The relations

\subsection{Model Selection and Optimization}
The core objective of graph learning procedures is selecting and training models that will be used for the downstream tasks, i.e., selecting and optimizing $\mathcal{A}$ based on the given data $\mathcal{G}$ and tasks $\mathcal{T}$. 
%The challenges are selecting and revising algorithms given the data and learning problem.
LLMs are indispensable in encoding the textual information and have advantages in making decisions due to the maintained knowledge.
%\footnote{+wln+ due to the large training data?}
In the following, we will introduce how LLMs are utilized given different types of data mentioned in Section~\ref{sec-feature}.
%, in which the involvements 
%and then introduce how LLMs are utilized when executing the algorithms.

%\footnote{+wln+ more methods can be shown in the table.}
\subsubsection{LLMs as Predictor for Textual Graph $\mathcal{G}^T$}
Existing methods employ open or closed source \textbf{LLMs}, denoted as $\mathcal{A}_{LLMs}$, to generate the prediction results $\mathcal{R}$ of graph learning problem $\mathcal{P}$.
%\footnote{+wln+ add equation here? \huan{yep, eq will be better, but try to make it simple}}
%i.e., $\mathcal{A}_{LLMs}: (\mathcal{G}^T, \mathcal{P}) \rightarrow \mathcal{R}$.
GraphText~\cite{zhao2023graphtext}, a representative method in the field, used closed-source ChatGPT to predict the node labels with k-hop sub-graph text in a few-shot manner.
TLG~\cite{fatemi2023talk} employed open-source PaLM 62B~\cite{chowdhery2023palm} to predict different graph properties via the text descriptions, which could be conducted in zero-shot manner.

In conclusion, when dealing with the textual graph, the utilization of LLMs is indispensable. 
These methods pose superior capabilities in generating explainable predictions compared with general graph learning methods such as GNNs and graph transformer~\cite{zhao2023graphtext}. 
Furthermore, they could used in zero-shot manner, which could adapt to different tasks and data easily due to the strong ability in understanding natural language queries.
Nonetheless, the prompt design and the usage of examples still have large influence on the model performance~\cite{fatemi2023talk}.
Moreover, the robustness, reliability, and data leakage problems of LLMs are the potential issues that need to be solved~\cite{huang2023can}.
%the results may suffer from the weakness of robustness and inreliable due to the data leakage~\cite{huang2023can}.
%\footnote{+wln+ fine-tuned on graph learning problems?}

\subsubsection{LLMs as Co-operator for Structured-Textual Graphs $\mathcal{G}^{ST}$}
\begin{figure}[ht]
	\centering
	\vspace{-5pt}
	\includegraphics[width=0.95\linewidth]{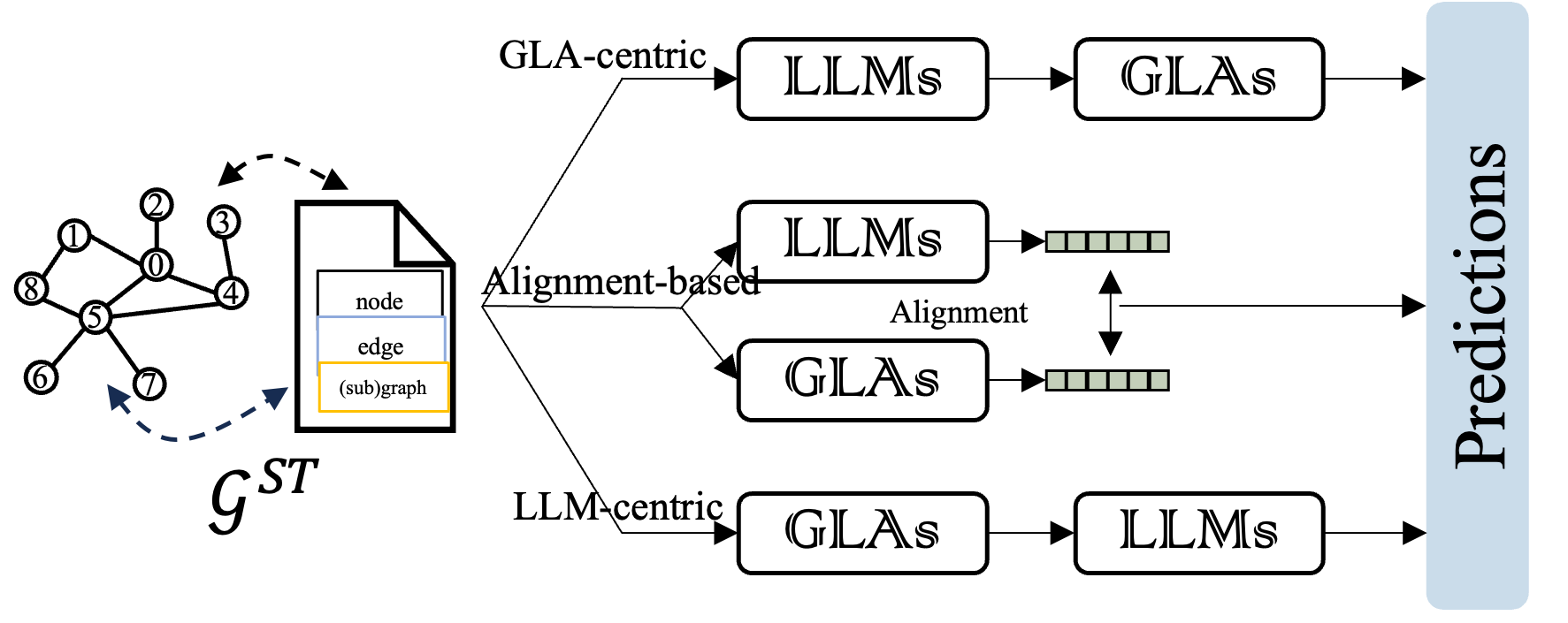}
	\vspace{-10pt}
	\caption{Illustrations of LLM-related graph learning algorithms on structured-textual graphs.}
		\vspace{-5pt}
	\label{fig-gla-gst}
\end{figure}

%When facing the graphs $G^{TA}$, 
With textual and structured graph, LLMs $\mathcal{A}_{LLMs}$ and graph learning algorithms $\mathcal{A}_{GLAs}$, like GNNs and graph Transformers~\cite{kipf2016semi,min2022transformer,wu2020comprehensive,wang2023automated},
%\huan{Refs for these two terms, could include work of wangxu:-)}, 
are co-operated towards accurate predictions.
% and this procedure can be denoted as$(\mathcal{A}_{GNN},\mathcal{A}_{LLMs}): (\mathcal{G}^{ST}, \mathcal{P})\rightarrow \mathcal{R}$.
Existing methods are organized following three categories as illustrated in Fig.~\ref{fig-gla-gst}.
Firstly, the \textbf{GLA-centric} methods first use LLMs to encode the textual information, and then employ graph learning algorithms to obtain predictions based on these enriched features.
By encoding the textual information of nodes and edges with Sentence-BERT firstly, OFA~\cite{liu2023one} adopted R-GCN~\cite{schlichtkrull2018modeling} to aggregate the features given different types of edges.
TAPE~\cite{he2023explanations} adopted Deberta to encode the textual descriptions and explanations , and then used RevGAT to learning graph representations.
%With (L)LMs, the text embedding could be pre-calculated when constructing graphs, and then they can be used by GNNs directly or aligned with the structured embedding. 
Secondly, the \textbf{alignment-based} methods are co-operated by aligning the textual and structured embedding space.
For instance, G2P2~\cite{wen2023prompt} adopted three levels contrastive learning when pre-trained the GCN~\cite{kipf2016semi} and Transformer~\cite{vaswani2017attention}.
Patton~\cite{jin2023patton} trained GNNs and BERT in a nested manner, and GRAD~\cite{mavromatis2023train} distilled the structured information from GNNs to enhance BERT towards better understanding on graphs.
Finally, the \textbf{LLM-centric} methods use graph learning algorithms to extract the graph structure information,
and then mapped into the text space of LLMs.
These methods could leverage the advantages of graph learning algorithms in understanding the inherent structural characteristics and LLMs in feature transformation~\cite{li2023survey}.
The representative GraphLLM~\cite{chai2023graphllm} first learned graph representations with Graph Transformer, and then mapped into prefix and tuned with LLaMA.

%Then, graph learning algorithms, like GNNs and graph Transformers, can be used merely to encode the graph structure information. Based on these methods, LLMs are treated as the graph enhancers in graph learning procedures~\cite{chen2023exploring,li2023survey}. 

%For instance, with the encoded features of nodes and edges, OFA~\cite{liu2023one} adopted R-GCN~\cite{schlichtkrull2018modeling} to aggregate the features given different types of edges. 
%\footnote{+wln+ it adopt with frozen LLMs and ft LMs }
%Besides, the frozen LLMs can be utilized aligned to train the graph learning methods. 

%To further exploring the encoding ability, graph learning models can be trained with the assist of encoded features (alignment-contrastive ). 
%\footnote{+wln+ gnn+llm alignment should be the training strategy? rather than model selection strategy?}
%Secondly, GNNs and open-sourced LLMs can be trained together towards better performance. 

In conclusion, LLMs and graph learning algorithms are co-operated when facing the structured-textual graphs.
Three categories of combinations manners are explored in the literature, i.e., GLA-centric, alignment-based and LLM-centric methods, as shown in Fig.~\ref{fig-gla-gst}. 
In these methods, LLMs are treated as an pure large models or as language models to enhancing the information extraction, which can be frozen or trained with GNNs.
%There is no certain evidence on 
In the future work, the improvements can be developed towards larger graph learning models given the model size of LLMs.
%and the former played an indispensable role when integrated the text modality. With the enriched information from this new modality, potential higher performance are expected. 
%However, how to well integrate these new modalities and how to compensate the disadvantages of LLMs is still unknown.

%Apart from the ability of encoding the text information, several methods employ LLMs to design algorithms via its strong reasoning ability and human-like intelligence.
%i.e., using LLMs as the core of autonomous agents to accomplish complex tasks.

\subsubsection{LLMs as Advisor for Structured graph $\mathcal{G}^S$}
Based on the structured graphs without textual information, 
%LLMs are utilized to design algorithms
the \textbf{LLM-assisted graph learning} methods, i.e., LLMs serve as the graph learning research assist in selecting models, are proposed considering its strong ability in reasoning and human-like decision making ability.
%rather than the ability of encoding the text information.
GPT4GNAS~\cite{wang2023graph} used GPT-4 to guide the designing of GNNs with AutoML. 
It used a fixed search space and then let GPT-4 select and revise GNNs from this space towards better-performed architectures. 
Instruction2GL~\cite{wei2023unleashing} employs GPT-3.5 to configure the search space and 
search algorithms to automated conducting graph learning procedures based on the AutoML technique. 
%These methods have advantages in obtaining flexible and explainable results, but may suffer from the stability and robustness issues due to the usage of LLMs.

\begin{figure}[ht]
	\centering
		\vspace{-5pt}
	\includegraphics[width=0.8\linewidth]{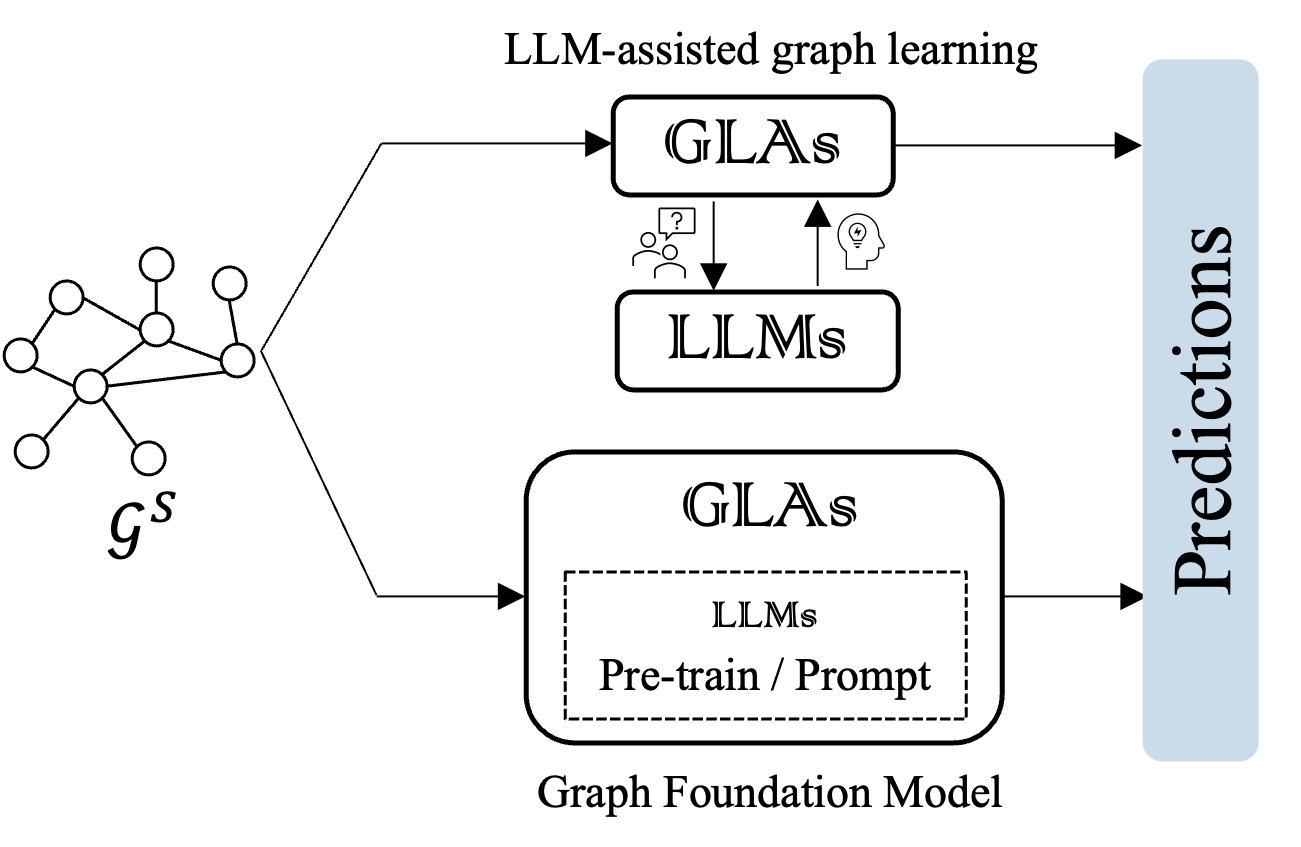}
	\vspace{-10pt}
	\caption{Illustrations of LLM-empowered graph learning algorithms on graph-structured data.}
	\label{fig-gla-gs}
	\vspace{-5pt}
\end{figure}

Moreover, motivated by the advantages of LLMs
%language foundation models, i.e., LLMs, 
in handling different tasks by training on large-scale data from different tasks, the development of \textbf{graph foundation models (GFMs)} are popular in recent years.
%\footnote{+wln+ is it proper to say the language foundation models as LLMs?}
%for which employ LLMs in an implicit manner compared with the aforementioned methods.
%\footnote{+wln+ following the backbone pre-train / post-processing(prompt/fine-tune) pipeline}
%\footnote{+wln+ pre-train and ft paradigm have been discussed in~\cite{liu2023towards}. ft have been utilized in GNNs, while prompt not }
To be specific, 
the constructions of foundation models following the topics of architecture backbone, pre-training and adaption to downstream tasks~\cite{liu2023towards,zhang2023large,jin2023large}.
Motivated by the ``pre-train and prompt'' large model train paradigm, 
existing methods construct pre-training data based on different tasks and then propose graph prompt strategy towards versatile graph learning models.
%adopted the graph prompt methods along with the pre-training 
%graph prompting methods are proposed to unifying different graph learning tasks based on the GNNs. 
The representative methods GraphPrompt~\cite{liu2023graphprompt} and GPPT~\cite{sun2022gppt} unified the graph learning tasks on node, edge and graph levels with link prediction task and then pre-train the GNNs on different tasks. 
In the prompt tuning stage, GraphPrompt learned the prompt vector and GPPT designed prompt function on tasks and structures when facing the different downstream tasks.

%The training paradigms and construction of backbones of LLMs can be adopted in graph learning models.

In conclusion, when facing the structured graph, LLMs could serve as advisor to support the graph learning as shown in Fig.~\ref{fig-gla-gs}, i.e., serve as an research assistant with the human-like decision-making capabilities, and motivate the construction and training of graph foundation models. 
Compared with the latter, the LLM-assisted methods necessitate a higher level of LLM proficiency in human-like intelligence.
All these methods underscore the potential of LLMs to accomplish the complex graph learning tasks,
%which is a new trend and pivotal technique in constructing user-friendly and low-threshold graph learning techniques~\cite{wang2023survey,wei2023unleashing}.
marking a new trend and a critical technique in the development of user-friendly, accessible graph learning methods~\cite{wang2023survey,wei2023unleashing}.
%have more versatile roles and not limited to encoding the textual information. 
%in the future work, the LLM-based autonomous agents could be a new

%\begin{figure}[ht]
%	\centering
%	\includegraphics[width=0.8\linewidth]{fig/gla_gs}
%	\caption{Illustrations of LLM-empowered graph learning algorithms.}
%	\label{fig-algo-selection}
%\end{figure}

%\footnote{+wln+ graph prompted methods, i.e., graph foundation model should be analyzed additionally.}

%In conclusion, existing methods have explored how to use LLMs in encoding the textual information and how to 

\subsubsection{Summary and Discussions}
In graph learning model selection and optimization procedures, LLMs play different roles depending on the type of graph data encountered. 
They are treated as the predictor when facing the textual graph, and serve as the co-operator towards better combination from two modalities in structured-textual graphs. 
When facing these graph graphs, LLMs are explored due to its ability in understanding the text modality and transforming features. 
When facing the structured graph, LLMs are treated as the advisor to assist the graph learning by leveraging their human-level intelligence and utilizing innovative model training strategies.
%due to their human-level intelligence and the model training strategies.
Expected for the benefits by incorporating new modalities,  LLMs have difficulties in understanding the graph structures~\cite{fatemi2023talk,guo2023gpt4graph}.
Such deficiencies accumulate over time, contributing to weak robustness and a lack of theoretical justifications in graph learning tasks.
%on which the deficiencies are accumulated and leads to the weak robustness and the lack of theoretical justifications in graph learning tasks.
%It is need to be solved that how to unleashing the power of LLMs by integrated with the existing graph leaning methods which have better understanding on the graph structural information.
A crucial challenge that needs to be addressed involves harnessing the power of LLMs by integrating them with existing graph learning methods, which have a superior understanding of graph structural information.

\subsection{Deployment and Serving}
Given the user request, deploying solutions and then generating response for users are the final step of the graph learning pipeline, and they are tightly connected with users, which indicating that more human engagements are required to improve the usability and accessibility~\cite{yao2018taking}.
LLMs could generate codes based on the users requirements and call proper APIs related to the data~\cite{wang2023survey,huang2023benchmarking}.
Following this paradigm, LLMs are used as general AI research assistant in writing files, executing code and inspecting outputs~\cite{huang2023benchmarking,mialon2023gaia,zhang2023automl}, displaying high usability and interpretability in conducting the graph learning procedures.
On graph-structured data, the deployment and serving are related to the tools and packages they used. Graph Tool-former~\cite{zhang2023graph} used external API tools to load graph data, and generate prediction as well as explanations on graphs under different tasks. 
Instruction2GL~\cite{wei2023unleashing} learned to call APIs in Pytorch Geometric package to load data, execute graph learning code, and generate the response based on the users' requests and code execution logs.
As a conclusion, 
%these are the final machine learning procedures and are related to the external graph development packages and APIs.
LLMs serve as development assistants, with human-like ability to use various tools.
%and act as new interface due to the ability in generating natural language response. 
They also act as a new interface due to their ability to generate natural language responses. 
In the future, LLMs could play even more vital role after familiar with the external graph developments tools. 
This advancement will make graph-based machine learning more accessible and easy to use, thereby lowering the barriers to entry.

\subsection{Summary and Discussions}
In this section, we have provided detailed analysis on how LLMs are involved in different graph learning procedures. 
%and the detailed comparisons have shown in Table~\ref{tb-overview}.
%LLMs can handle the graph learning task definition procedure due to its ability in understanding the natural languages and the maintained knowledge about graphs~\cite{zhao2023survey,liu2023towards}.
%They can further construct and encode textual information, e.g., textual graphs and structured-textual graphs, to enhance the features~\cite{fatemi2023talk,he2023explanations}.
%Then, when selecting and optimizing the graph learning models on top of different graphs, LLMs serve as predictor, co-operator or advisor towards well-performed and versatile methods.
%Finally, LLMs assist the model deployment and response generation based on the results.
%To summarize, 
Given the versatile abilities of LLMs, it is feasible to integrate them towards versatile graph learning algorithms as shown in Fig.~\ref{fig-overview} and  Table~\ref{tb-overview}.
%while the graph learning procedures are well-established, the potential of LLMs applied on the procedures remains under-researched. 
%Given the fast developments of LLMs and the under-explored abilities of LLMs,
%there is still significant work to be done in considering how to incorporate LLMs into graph learning problems effectively.
Considering the rapid progression and yet under-explored potential of LLMs, significant research opportunities still exist to explore how to effectively integrate LLMs into graph learning problems.

\section{Future Directions}
The developments of LLMs and the potential applications are fast growing in recent years. 
Beyond existing methods, there are still several promising directions that could be explored on the graph-structured data.

\noindent\textbf{The ability quantification of LLMs in understanding graphs-structured data.}  
There is an absence of systematic evaluations regarding the ability of LLMs to comprehend graph-structured data. 
To be specific, LLMs are trained with natural languages, 
%and the understanding abilities are largely relied on the data they used.
while the graph structure, the key in graph learning, are represented as the adjacency matrix in general and are new modality for LLMs.
%Existing methods have evaluated that LLMs have deficiencies in understanding the graph structures~\cite{guo2023gpt4graph,fatemi2023talk,huang2023can}. 
Existing methods, e.g., GPT4Graph~\cite{guo2023gpt4graph} and TLG~\cite{fatemi2023talk}, attempt to describe graph structures with natural languages and then evaluate the effectiveness on graph structure reasoning tasks,
%GPT4Graph~\cite{guo2023gpt4graph} and TLG~\cite{fatemi2023talk} have provided several graph structure reasoning tasks to empirically evaluating the results, 
e.g., recognizing the node degree, graph diameter or clustering coefficient. 
However, these measures are not adequate to entirely capture the capacity of LLMs to understand graph structures. 
Firstly, the descriptions can greatly impact performance, which may result in insufficient robustness and reliability.
%Besides, Weisfeiler-Lehman test is the milestone when evaluating the effectiveness in understanding the graph structures, which would be considered to theoretically justifying the effectiveness.
Moreover, Weisfeiler-Lehman test, which is seminal in assessing the comprehension of graph structures, should be considered to theoretically validate the effectiveness. 
%Even with different input designing strategies and the prompt strategies, the results are far away compared with human-level (which assumed 100\% accuracy).
%Therefore, both from the theoretical and empirical perspective, there lacks a benchmark and detailed analysis to evaluating the ability of LLMs in understanding the graph structures.
Consequently, both from the theoretical and empirical points of view, there is a need for a benchmark and detailed analysis to evaluate the competence of LLMs in understanding graph structures.
% compared with human preferences in different levels, e.g., students in different degrees.
%One step further, Weisfeiler-Lehman test is the milestone when evaluating the effectiveness in understanding the graph structures, which would be considered to theoretically justifying the effectiveness.

%How to re-train or prompt tuning with graph-structure data is still unexplored.

%\noindent\textbf{The expressiveness of LLMs.}
%WL test is used to theoretically evaluating the graph learning problems. However, current LLMs even cannot understanding the graph structures. It is still a long distance to identifying the expressiveness of LLMs. New metric? or benchmark? (How to understand LLM expressiveness in NLP?)

%\noindent\textbf{LLM enhanced few-text/info graph augmentation?}

\noindent\textbf{Large graph foundation models.} 
Excepted for employing LLMs in solving graph learning tasks, designing graph foundation models and scaling up the model parameters is a promising direction in handling the different tasks and data from different domains.
To be specific, existing methods have explored the GNNs and Graph Transformer methods on the given single task~\cite{wu2020comprehensive,min2022transformer}.
%Compared with the foundation models in natural language, the challenges in designing the large graph model are
%the lack of extensive pre-train data and the large-scale backbone architectures
Compared with foundation models in natural language, challenges in designing large graph models are the scarcity of extensive pre-training data and the lack of large-scale backbone architectures
~\cite{liu2023towards,zhang2023large}.
Furthermore, designing the graph foundation models for specific graph domains is more effective and feasible considering the application scenarios~\cite{zhang2023large}, e.g., recommendation systems, molecules and finance.
%The data used to train foundation model, training strategies, backbone architectures \cite{liu2023towards,zhang2023large}. ( It is feasible to design for specific domains rather than general graphs)

%\noindent\textbf{Robustness and Interpretability} The deficiencies of LLMs.

\noindent\textbf{Universal Graph learning agents.} 
%Different with directly addressing the graph learning problems by human experts, LLMs have the potential serve as the core of autonomous agents. 
LLM-based autonomous agents are the prominent research topic in both academic and industry due to its potential emulating human-like decision-making processes~\cite{wang2023survey}.
In graph learning, LLMs could serve as the research assistant when facing the different learning problems.
For example, decomposing the graph learning tasks, writing and executing graph learning procedures step by step~\cite{mialon2023gaia,wei2023unleashing,huang2023benchmarking}, or arranging the graph learning tools when facing the tasks.
%In this way, the current graph learning techniques can be benefited.
Consequently, such an approach holds immense potential to facilitate advancements in current graph learning methods.
%When aching this, the challenges are relied on the 
%It could incorporating the current graph learning tools to solving the graph learning problems,
%Incorporating more graph learning \textbf{tools} and graph reasoning APIs to understand graph structures and semantic information,
%e.g., graph analysis metric and GL methods (networkx: shortest-path / node2vec / GNN)
%It is easier to achieve this compared with improving the basic ability of LLMs.

%\noindent\textbf{More benefits from LLMs to enhancing the graph learning.} More graphs may not be text-rich graphs, e.g., nodes in spatial-temporal traffic graphs. how to benefits all graphs with new modalities (text video, image or others?) These information may collected from internet or generated by LLMs.

%\noindent\textbf{Integrating graph learning knowledge with LLMs} Learn to understand graphs in higher-level rather than focus on encoding the given graphs. e.g., the developments of graph learning techniques and etc.

%\huan{How about a subsection to discuss versatile graph learning v.s. automated graph learning, since we have work on this area for more than 4 years. Besides, we can discuss this topic from the perspective of practical experiences, say, we have so many real-world applications in 4Paradigm, and we may kind of provide how versatile graph learning work in production and help business in 4Paradigm, which will show unique values provided by 4Paradigm researchers.}
%\footnote{+wln+ update in next version}

\section{Conclusion}
Exploring the feasibility of employing LLMs in graph learning problems is a promising direction considering the applications of graphs in the real world.
In this paper, we propose a novel conceptual prototype for versatile graph learning approaches, and provide a comprehensive overview of existing methods jointed with LLMs following the graph learning procedures and the different ability requirements for LLMs.
%and give systematical taxonomy following the graph learning procedures and the involvements of LLMs.
To be specific, we first explore the feasibility of LLMs used in the graph learning pipeline, including task definition, graph feature engineering, model selection and optimization, deployment and serving procedures. 
In each procedure, we categorize methods following the involvements of LLMs and provide detailed discussions on these methods.
Finally, we suggest future directions for LLM-based graph learning approaches, potentially expanding the methodology towards universal graph learning methods.
%\footnote{+wln+ no need to introduce the procedures and methods in detail. the advantages and benefits.}

\clearpage
\appendix

%\section*{Ethical Statement}
%
%There are no ethical issues.

\clearpage
%\section*{Acknowledgments}

%The preparation of these instructions and the \LaTeX{} and Bib\TeX{}
%files that implement them was supported by Schlumberger Palo Alto
%Research, AT\&T Bell Laboratories, and Morgan Kaufmann Publishers.
%Preparation of the Microsoft Word file was supported by IJCAI.  An
%early version of this document was created by Shirley Jowell and Peter
%F. Patel-Schneider.  It was subsequently modified by Jennifer
%Ballentine, Thomas Dean, Bernhard Nebel, Daniel Pagenstecher,
%Kurt Steinkraus, Toby Walsh, Carles Sierra, Marc Pujol-Gonzalez,
%Francisco Cruz-Mencia and Edith Elkind.

%% The file named.bst is a bibliography style file for BibTeX 0.99c
\bibliographystyle{named}
\bibliography{ijcai24}

\begin{thebibliography}{}

\bibitem[\protect\citeauthoryear{Chai \bgroup \em et al.\egroup
  }{2023}]{chai2023graphllm}
Ziwei Chai, Tianjie Zhang, Liang Wu, Kaiqiao Han, Xiaohai Hu, Xuanwen Huang,
  and Yang Yang.
\newblock Graphllm: Boosting graph reasoning ability of large language model.
\newblock {\em arXiv preprint arXiv:2310.05845}, 2023.

\bibitem[\protect\citeauthoryear{Chen and Bertozzi}{2023}]{chen2023autokg}
Bohan Chen and Andrea~L Bertozzi.
\newblock Autokg: Efficient automated knowledge graph generation for language
  models.
\newblock {\em arXiv preprint arXiv:2311.14740}, 2023.

\bibitem[\protect\citeauthoryear{Chen \bgroup \em et al.\egroup
  }{2023}]{chen2023exploring}
Zhikai Chen, Haitao Mao, Hang Li, Wei Jin, Hongzhi Wen, Xiaochi Wei, Shuaiqiang
  Wang, Dawei Yin, Wenqi Fan, Hui Liu, et~al.
\newblock Exploring the potential of large language models (llms) in learning
  on graphs.
\newblock {\em arXiv preprint arXiv:2307.03393}, 2023.

\bibitem[\protect\citeauthoryear{Chowdhery \bgroup \em et al.\egroup
  }{2023}]{chowdhery2023palm}
Aakanksha Chowdhery, Sharan Narang, Jacob Devlin, Maarten Bosma, Gaurav Mishra,
  Adam Roberts, Paul Barham, Hyung~Won Chung, Charles Sutton, Sebastian
  Gehrmann, et~al.
\newblock Palm: Scaling language modeling with pathways.
\newblock {\em Journal of Machine Learning Research}, 24(240):1--113, 2023.

\bibitem[\protect\citeauthoryear{Fatemi \bgroup \em et al.\egroup
  }{2023}]{fatemi2023talk}
Bahare Fatemi, Jonathan Halcrow, and Bryan Perozzi.
\newblock Talk like a graph: Encoding graphs for large language models.
\newblock {\em arXiv preprint arXiv:2310.04560}, 2023.

\bibitem[\protect\citeauthoryear{Ge \bgroup \em et al.\egroup
  }{2023}]{ge2023openagi}
Yingqiang Ge, Wenyue Hua, Jianchao Ji, Juntao Tan, Shuyuan Xu, and Yongfeng
  Zhang.
\newblock Openagi: When llm meets domain experts.
\newblock {\em arXiv preprint arXiv:2304.04370}, 2023.

\bibitem[\protect\citeauthoryear{Gilmer \bgroup \em et al.\egroup
  }{2017}]{gilmer2017neural}
Justin Gilmer, Samuel~S Schoenholz, Patrick~F Riley, Oriol Vinyals, and
  George~E Dahl.
\newblock Neural message passing for quantum chemistry.
\newblock In {\em ICML}, pages 1263--1272, 2017.

\bibitem[\protect\citeauthoryear{Guo \bgroup \em et al.\egroup
  }{2023}]{guo2023gpt4graph}
Jiayan Guo, Lun Du, and Hengyu Liu.
\newblock Gpt4graph: Can large language models understand graph structured
  data? an empirical evaluation and benchmarking.
\newblock {\em arXiv preprint arXiv:2305.15066}, 2023.

\bibitem[\protect\citeauthoryear{Hamilton \bgroup \em et al.\egroup
  }{2017}]{hamilton2017inductive}
Will Hamilton, Zhitao Ying, and Jure Leskovec.
\newblock Inductive representation learning on large graphs.
\newblock In {\em NeurIPS}, pages 1024--1034, 2017.

\bibitem[\protect\citeauthoryear{He \bgroup \em et al.\egroup
  }{2023}]{he2023explanations}
Xiaoxin He, Xavier Bresson, Thomas Laurent, and Bryan Hooi.
\newblock Explanations as features: Llm-based features for text-attributed
  graphs.
\newblock {\em arXiv preprint arXiv:2305.19523}, 2023.

\bibitem[\protect\citeauthoryear{Huang \bgroup \em et al.\egroup
  }{2023a}]{huang2023can}
Jin Huang, Xingjian Zhang, Qiaozhu Mei, and Jiaqi Ma.
\newblock Can llms effectively leverage graph structural information: when and
  why.
\newblock {\em arXiv preprint arXiv:2309.16595}, 2023.

\bibitem[\protect\citeauthoryear{Huang \bgroup \em et al.\egroup
  }{2023b}]{huang2023benchmarking}
Qian Huang, Jian Vora, Percy Liang, and Jure Leskovec.
\newblock Benchmarking large language models as ai research agents.
\newblock {\em arXiv preprint arXiv:2310.03302}, 2023.

\bibitem[\protect\citeauthoryear{Jin \bgroup \em et al.\egroup
  }{2023a}]{jin2023large}
Bowen Jin, Gang Liu, Chi Han, Meng Jiang, Heng Ji, and Jiawei Han.
\newblock Large language models on graphs: A comprehensive survey.
\newblock {\em arXiv preprint arXiv:2312.02783}, 2023.

\bibitem[\protect\citeauthoryear{Jin \bgroup \em et al.\egroup
  }{2023b}]{jin2023patton}
Bowen Jin, Wentao Zhang, Yu~Zhang, Yu~Meng, Xinyang Zhang, Qi~Zhu, and Jiawei
  Han.
\newblock Patton: Language model pretraining on text-rich networks.
\newblock {\em arXiv preprint arXiv:2305.12268}, 2023.

\bibitem[\protect\citeauthoryear{Kipf and Welling}{2016}]{kipf2016semi}
Thomas~N Kipf and Max Welling.
\newblock Semi-supervised classification with graph convolutional networks.
\newblock {\em ICLR}, 2016.

\bibitem[\protect\citeauthoryear{Li \bgroup \em et al.\egroup
  }{2023}]{li2023survey}
Yuhan Li, Zhixun Li, Peisong Wang, Jia Li, Xiangguo Sun, Hong Cheng, and
  Jeffrey~Xu Yu.
\newblock A survey of graph meets large language model: Progress and future
  directions.
\newblock {\em arXiv preprint arXiv:2311.12399}, 2023.

\bibitem[\protect\citeauthoryear{Liu \bgroup \em et al.\egroup
  }{2019}]{liu2019roberta}
Yinhan Liu, Myle Ott, Naman Goyal, Jingfei Du, Mandar Joshi, Danqi Chen, Omer
  Levy, Mike Lewis, Luke Zettlemoyer, and Veselin Stoyanov.
\newblock Roberta: A robustly optimized bert pretraining approach.
\newblock {\em arXiv preprint arXiv:1907.11692}, 2019.

\bibitem[\protect\citeauthoryear{Liu \bgroup \em et al.\egroup
  }{2022}]{liu2022taxonomy}
Renming Liu, Semih Cant{\"u}rk, Frederik Wenkel, Sarah McGuire, Xinyi Wang,
  Anna Little, Leslie O’Bray, Michael Perlmutter, Bastian Rieck, Matthew
  Hirn, et~al.
\newblock Taxonomy of benchmarks in graph representation learning.
\newblock In {\em Learning on Graphs Conference}, pages 6--1. PMLR, 2022.

\bibitem[\protect\citeauthoryear{Liu \bgroup \em et al.\egroup
  }{2023a}]{liu2023one}
Hao Liu, Jiarui Feng, Lecheng Kong, Ningyue Liang, Dacheng Tao, Yixin Chen, and
  Muhan Zhang.
\newblock One for all: Towards training one graph model for all classification
  tasks.
\newblock {\em arXiv preprint arXiv:2310.00149}, 2023.

\bibitem[\protect\citeauthoryear{Liu \bgroup \em et al.\egroup
  }{2023b}]{liu2023towards}
Jiawei Liu, Cheng Yang, Zhiyuan Lu, Junze Chen, Yibo Li, Mengmei Zhang, Ting
  Bai, Yuan Fang, Lichao Sun, Philip~S Yu, et~al.
\newblock Towards graph foundation models: A survey and beyond.
\newblock {\em arXiv preprint arXiv:2310.11829}, 2023.

\bibitem[\protect\citeauthoryear{Liu \bgroup \em et al.\egroup
  }{2023c}]{liu2023graphprompt}
Zemin Liu, Xingtong Yu, Yuan Fang, and Xinming Zhang.
\newblock Graphprompt: Unifying pre-training and downstream tasks for graph
  neural networks.
\newblock In {\em Proceedings of the ACM Web Conference 2023}, pages 417--428,
  2023.

\bibitem[\protect\citeauthoryear{Mavromatis \bgroup \em et al.\egroup
  }{2023}]{mavromatis2023train}
Costas Mavromatis, Vassilis~N Ioannidis, Shen Wang, Da~Zheng, Soji Adeshina,
  Jun Ma, Han Zhao, Christos Faloutsos, and George Karypis.
\newblock Train your own gnn teacher: Graph-aware distillation on textual
  graphs.
\newblock {\em arXiv preprint arXiv:2304.10668}, 2023.

\bibitem[\protect\citeauthoryear{Mialon \bgroup \em et al.\egroup
  }{2023}]{mialon2023gaia}
Gr{\'e}goire Mialon, Cl{\'e}mentine Fourrier, Craig Swift, Thomas Wolf, Yann
  LeCun, and Thomas Scialom.
\newblock Gaia: a benchmark for general ai assistants.
\newblock {\em arXiv preprint arXiv:2311.12983}, 2023.

\bibitem[\protect\citeauthoryear{Min \bgroup \em et al.\egroup
  }{2022}]{min2022transformer}
Erxue Min, Runfa Chen, Yatao Bian, Tingyang Xu, Kangfei Zhao, Wenbing Huang,
  Peilin Zhao, Junzhou Huang, Sophia Ananiadou, and Yu~Rong.
\newblock Transformer for graphs: An overview from architecture perspective.
\newblock {\em arXiv preprint arXiv:2202.08455}, 2022.

\bibitem[\protect\citeauthoryear{Qian \bgroup \em et al.\egroup
  }{2023}]{qian2023can}
Chen Qian, Huayi Tang, Zhirui Yang, Hong Liang, and Yong Liu.
\newblock Can large language models empower molecular property prediction?
\newblock {\em arXiv preprint arXiv:2307.07443}, 2023.

\bibitem[\protect\citeauthoryear{Reimers and
  Gurevych}{2019}]{reimers2019sentence}
Nils Reimers and Iryna Gurevych.
\newblock Sentence-bert: Sentence embeddings using siamese bert-networks.
\newblock {\em arXiv preprint arXiv:1908.10084}, 2019.

\bibitem[\protect\citeauthoryear{Schlichtkrull \bgroup \em et al.\egroup
  }{2018}]{schlichtkrull2018modeling}
Michael Schlichtkrull, Thomas~N Kipf, Peter Bloem, Rianne Van Den~Berg, Ivan
  Titov, and Max Welling.
\newblock Modeling relational data with graph convolutional networks.
\newblock In {\em The Semantic Web: 15th International Conference, ESWC 2018,
  Heraklion, Crete, Greece, June 3--7, 2018, Proceedings 15}, pages 593--607.
  Springer, 2018.

\bibitem[\protect\citeauthoryear{Shen \bgroup \em et al.\egroup
  }{2023}]{shen2023hugginggpt}
Yongliang Shen, Kaitao Song, Xu~Tan, Dongsheng Li, Weiming Lu, and Yueting
  Zhuang.
\newblock Hugginggpt: Solving ai tasks with chatgpt and its friends in
  huggingface.
\newblock {\em arXiv preprint arXiv:2303.17580}, 2023.

\bibitem[\protect\citeauthoryear{Sun \bgroup \em et al.\egroup
  }{2022}]{sun2022gppt}
Mingchen Sun, Kaixiong Zhou, Xin He, Ying Wang, and Xin Wang.
\newblock Gppt: Graph pre-training and prompt tuning to generalize graph neural
  networks.
\newblock In {\em Proceedings of the 28th ACM SIGKDD Conference on Knowledge
  Discovery and Data Mining}, pages 1717--1727, 2022.

\bibitem[\protect\citeauthoryear{Tang \bgroup \em et al.\egroup
  }{2023}]{tang2023graphgpt}
Jiabin Tang, Yuhao Yang, Wei Wei, Lei Shi, Lixin Su, Suqi Cheng, Dawei Yin, and
  Chao Huang.
\newblock Graphgpt: Graph instruction tuning for large language models.
\newblock {\em arXiv preprint arXiv:2310.13023}, 2023.

\bibitem[\protect\citeauthoryear{Vaswani \bgroup \em et al.\egroup
  }{2017}]{vaswani2017attention}
Ashish Vaswani, Noam Shazeer, Niki Parmar, Jakob Uszkoreit, Llion Jones,
  Aidan~N Gomez, {\L}ukasz Kaiser, and Illia Polosukhin.
\newblock Attention is all you need.
\newblock {\em Advances in neural information processing systems}, 30, 2017.

\bibitem[\protect\citeauthoryear{Wang \bgroup \em et al.\egroup
  }{2023a}]{wang2023graph}
Haishuai Wang, Yang Gao, Xin Zheng, Peng Zhang, Hongyang Chen, and Jiajun Bu.
\newblock Graph neural architecture search with gpt-4.
\newblock {\em arXiv preprint arXiv:2310.01436}, 2023.

\bibitem[\protect\citeauthoryear{Wang \bgroup \em et al.\egroup
  }{2023b}]{wang2023can}
Heng Wang, Shangbin Feng, Tianxing He, Zhaoxuan Tan, Xiaochuang Han, and Yulia
  Tsvetkov.
\newblock Can language models solve graph problems in natural language?
\newblock {\em arXiv preprint arXiv:2305.10037}, 2023.

\bibitem[\protect\citeauthoryear{Wang \bgroup \em et al.\egroup
  }{2023c}]{wang2023survey}
Lei Wang, Chen Ma, Xueyang Feng, Zeyu Zhang, Hao Yang, Jingsen Zhang, Zhiyuan
  Chen, Jiakai Tang, Xu~Chen, Yankai Lin, et~al.
\newblock A survey on large language model based autonomous agents.
\newblock {\em arXiv preprint arXiv:2308.11432}, 2023.

\bibitem[\protect\citeauthoryear{Wang \bgroup \em et al.\egroup
  }{2023d}]{wang2023automated}
Xu~Wang, Huan Zhao, Wei-wei Tu, and Quanming Yao.
\newblock Automated 3d pre-training for molecular property prediction.
\newblock In {\em Proceedings of the 29th ACM SIGKDD Conference on Knowledge
  Discovery and Data Mining}, pages 2419--2430, 2023.

\bibitem[\protect\citeauthoryear{Wei \bgroup \em et al.\egroup
  }{2022}]{wei2022chain}
Jason Wei, Xuezhi Wang, Dale Schuurmans, Maarten Bosma, Fei Xia, Ed~Chi, Quoc~V
  Le, Denny Zhou, et~al.
\newblock Chain-of-thought prompting elicits reasoning in large language
  models.
\newblock {\em Advances in Neural Information Processing Systems},
  35:24824--24837, 2022.

\bibitem[\protect\citeauthoryear{Wei \bgroup \em et al.\egroup
  }{2023}]{wei2023unleashing}
Lanning Wei, Zhiqiang He, Huan Zhao, and Quanming Yao.
\newblock Unleashing the power of graph learning through llm-based autonomous
  agents.
\newblock {\em arXiv preprint arXiv:2309.04565}, 2023.

\bibitem[\protect\citeauthoryear{Wen and Fang}{2023}]{wen2023prompt}
Zhihao Wen and Yuan Fang.
\newblock Prompt tuning on graph-augmented low-resource text classification.
\newblock {\em arXiv preprint arXiv:2307.10230}, 2023.

\bibitem[\protect\citeauthoryear{Wu \bgroup \em et al.\egroup
  }{2020}]{wu2020comprehensive}
Zonghan Wu, Shirui Pan, Fengwen Chen, Guodong Long, Chengqi Zhang, and S~Yu
  Philip.
\newblock A comprehensive survey on graph neural networks.
\newblock {\em IEEE Transactions on Neural Networks and Learning Systems
  (TNNLS)}, 2020.

\bibitem[\protect\citeauthoryear{Yao \bgroup \em et al.\egroup
  }{2018}]{yao2018taking}
Quanming Yao, Mengshuo Wang, Yuqiang Chen, Wenyuan Dai, Yu-Feng Li, Wei-Wei Tu,
  Qiang Yang, and Yang Yu.
\newblock Taking human out of learning applications: A survey on automated
  machine learning.
\newblock {\em arXiv preprint arXiv:1810.13306}, 2018.

\bibitem[\protect\citeauthoryear{You \bgroup \em et al.\egroup
  }{2020}]{you2020design}
Jiaxuan You, Zhitao Ying, and Jure Leskovec.
\newblock Design space for graph neural networks.
\newblock {\em Advances in Neural Information Processing Systems},
  33:17009--17021, 2020.

\bibitem[\protect\citeauthoryear{Zhang \bgroup \em et al.\egroup
  }{2021}]{zhang2021automated}
Ziwei Zhang, Xin Wang, and Wenwu Zhu.
\newblock Automated machine learning on graphs: A survey.
\newblock 2021.
\newblock Survey track.

\bibitem[\protect\citeauthoryear{Zhang \bgroup \em et al.\egroup
  }{2022}]{zhang2022efficient}
Yongqi Zhang, Zhanke Zhou, Quanming Yao, and Yong Li.
\newblock Efficient hyper-parameter search for knowledge graph embedding.
\newblock In {\em Proceedings of the 60th Annual Meeting of the Association for
  Computational Linguistics (Volume 1: Long Papers)}, pages 2715--2735, 2022.

\bibitem[\protect\citeauthoryear{Zhang \bgroup \em et al.\egroup
  }{2023a}]{zhang2023automl}
Shujian Zhang, Chengyue Gong, Lemeng Wu, Xingchao Liu, and Mingyuan Zhou.
\newblock Automl-gpt: Automatic machine learning with gpt.
\newblock {\em arXiv preprint arXiv:2305.02499}, 2023.

\bibitem[\protect\citeauthoryear{Zhang \bgroup \em et al.\egroup
  }{2023b}]{zhang2023large}
Ziwei Zhang, Haoyang Li, Zeyang Zhang, Yijian Qin, Xin Wang, and Wenwu Zhu.
\newblock Large graph models: A perspective.
\newblock {\em arXiv preprint arXiv:2308.14522}, 2023.

\bibitem[\protect\citeauthoryear{Zhang}{2023}]{zhang2023graph}
Jiawei Zhang.
\newblock Graph-toolformer: To empower llms with graph reasoning ability via
  prompt augmented by chatgpt.
\newblock {\em arXiv preprint arXiv:2304.11116}, 2023.

\bibitem[\protect\citeauthoryear{Zhao \bgroup \em et al.\egroup
  }{2023a}]{zhao2023graphtext}
Jianan Zhao, Le~Zhuo, Yikang Shen, Meng Qu, Kai Liu, Michael Bronstein,
  Zhaocheng Zhu, and Jian Tang.
\newblock Graphtext: Graph reasoning in text space.
\newblock {\em arXiv preprint arXiv:2310.01089}, 2023.

\bibitem[\protect\citeauthoryear{Zhao \bgroup \em et al.\egroup
  }{2023b}]{zhao2023survey}
Wayne~Xin Zhao, Kun Zhou, Junyi Li, Tianyi Tang, Xiaolei Wang, Yupeng Hou,
  Yingqian Min, Beichen Zhang, Junjie Zhang, Zican Dong, et~al.
\newblock A survey of large language models.
\newblock {\em arXiv preprint arXiv:2303.18223}, 2023.

\bibitem[\protect\citeauthoryear{Zheng \bgroup \em et al.\egroup
  }{2023}]{zheng2023automl}
Ruiqi Zheng, Liang Qu, Bin Cui, Yuhui Shi, and Hongzhi Yin.
\newblock Automl for deep recommender systems: A survey.
\newblock {\em ACM Transactions on Information Systems}, 41(4):1--38, 2023.

\end{thebibliography}

\end{document}